\documentclass[11pt]{article}
\usepackage{coling2018}
\usepackage[hidelinks]{hyperref}
\usepackage{times}
\usepackage{latexsym}
\usepackage{amsmath}
\usepackage{amssymb}
\usepackage{microtype}
\usepackage[utf8]{inputenc}
\usepackage{paralist}
\usepackage{nicefrac}
\usepackage{textcomp} 
\usepackage{booktabs}
\usepackage{array}
\usepackage{graphicx}
\usepackage{tikz}
\usetikzlibrary{calc}
\usetikzlibrary{shapes}
\usetikzlibrary{matrix}
\usetikzlibrary{arrows}
\usetikzlibrary{decorations.pathreplacing}
\usetikzlibrary{positioning}
\usetikzlibrary{backgrounds}
\usetikzlibrary{shadows.blur}

\usepackage[ngerman,british]{babel}

\usepackage{relsize}
\usepackage{tipa,extraipa}
\usepackage{wrapfig}
\usepackage[disable]{todonotes}

\slname{German}

\title{Incremental Natural Language Processing:\\ Challenges, Strategies, and Evaluation}

\sltitle{Inkrementelle Sprachverarbeitung:\\ Herausforderungen, Strategien und Evaluation}

\newcommand{\correct}[1]{{\color{greentextuhh}#1}}
\newcommand{\incorrect}[1]{{\color{reduhh}#1}}

\newcommand\ncoord[2][0,0]{%
    \tikz[remember picture,overlay]{\path (#1) coordinate (#2);}%

}
\definecolor{blueuhh}{RGB}{0,156,209}
\definecolor{reduhh}{RGB}{226,0,26}
\definecolor{greyuhh}{RGB}{59,81,91}
\definecolor{lightgreyuhh}{RGB}{156,167,172}
\definecolor{bluetextuhh}{RGB}{0,133,179}
\definecolor{greentextuhh}{RGB}{69,178,72}

\author{Arne Köhn\\
  Natural Lanuguage Systems Group\\
  Department of Informatics\\
  Universität Hamburg\\
  \texttt{\href{mailto:koehn@informatik.uni-hamburg.de}{koehn@informatik.uni-hamburg.de}}\\
}

\hypersetup{
pdfinfo={
Title={Incremental Natural Language Processing: Challenges, Strategies, and Evaluation},
Author={Arne Köhn},
% ...
}
}

\begin{document}
\maketitle
\begin{abstract}
Incrementality is ubiquitous in human-human interaction and beneficial
for human-computer interaction.  It has been a topic of research in
different parts of the NLP community, mostly with focus on the
specific topic at hand even though incremental systems have to deal
with similar challenges regardless of domain.
In this survey,
I consolidate and categorize the approaches, identifying
similarities and differences in the computation and data, and show
trade-offs that have to be considered.  A
focus lies on evaluating incremental systems because
the standard metrics
often fail to capture the incremental properties of a system and coming up with
a suitable evaluation scheme is non-trivial.
\end{abstract}

\makesltitle
\begin{slabstract}
  \begin{otherlanguage}{ngerman}
Inkrementalität ist allgegenwärtig in Mensch-Mensch-Interaktiton
und hilfreich für Mensch-Computer-Interaktion.  In verschiedenen
Teilen der NLP-Community wird an Inkrementalität geforscht,
zumeist fokussiert auf eine konkrete Aufgabe, obwohl sich inkrementellen
Systemen domänenübergreifend ähnliche Herausforderungen stellen.
In diesem Überblick trage ich Ansätze zusammen, kategorisiere
sie und stelle Ähnlichkeiten und Unterschiede in Berechnung
und Daten sowie nötige Abwägungen vor.  Ein Fokus liegt auf
der Evaluierung inkrementeller Systeme, da Standardmetriken of
nicht in der Lage sind, die inkrementellen Eigenschaften eines
Systems einzufangen und passende Evaluationsschemata zu entwickeln
nicht einfach ist.
  \end{otherlanguage}
\end{slabstract}

\section{Introduction}
\label{sec:intro}
\blfootnote{
  \hspace{-0.72cm}  
  This work is licensed under a Creative Commons 
  Attribution 4.0 International License.
  License details:
  \url{http://creativecommons.org/licenses/by/4.0/}.
}

Interaction using language is incremental in many forms: In a dialogue,
understanding takes place continuously and participants are able to
interrupt each other or signal whether they understand the currently ongoing
utterance.  Simultaneous interpreters translate speeches as they are
spoken.  Texts are (partially) understood before they are fully read.
%
Incremental processing exploits this property by starting to
compute before all input is available, allowing a system to already
act on partial input.
Without incremental processing, an NLP system is unable to perform any
of these tasks as it is bound to wait for complete utterances or a
finished text and only afterwards it can compute how to (re-)act, leading
to deficiencies in human-computer interaction.

First, I want to delimit this notion of incrementality from other notions:
Systems that work on a complete input but generate output
layer by layer, e.g.\ shallow to deep syntax, are sometimes called
incremental, e.g.\ by \newcite{ait-mokhtar-2002-robustnes-inc-deep-parsing}.  These
systems are not discussed in this survey.
Anytime algorithms are incremental in the sense that
they iteratively improve their
output for a fixed input, given more and more processing time. They can play an
important role in incremental systems as they allow to perform a
trade-off between processing time -- i.e.\
system responsiveness -- and quality.  The question of whether
to employ anytime algorithms is however orthogonal to the approaches
discussed in this survey\footnote{An anytime algorithm could be seen
  as a simply another non-monotonic processor, as described in Section \ref{sec:monotonicity}.}.

In an incremental system, all processors need to work
incrementally.  A prime example is the Verbmobil project, which set out
to develop a portable simultaneous interpreter
\cite{kay-1994-verbmobil}. The project developed speech recognition
and synthesis components, syntactic and semantic parsers,
self-correction detection, dialogue modeling and of course machine
translation, showing that incrementality is an aspect that touches nearly all topics
of NLP.
This project also exemplifies that building incremental
systems is not easy, even with massive funding\footnote{Verbmobil had a
  funding of 116 million DM, (\texttildelow 60 million €, or 78 million € when adjusted for inflation).}:
Only one of the many components ended up being incremental and the
final report makes no mention of \emph{simultaneous} interpretation
\cite{wahlster-2000-verbmobil}.

The incremental nature is more obvious in speech than in written
language because language processing often processes
already-written text in bulk whereas interactive systems are a major
research focus for speech-based applications.
As the benefits of incrementality are more prominent in processing
dialogues compared to text, research on incrementality in NLP has been primarily
driven by speech-based research questions,
such as
understanding based on partial speech recognition \cite{sagae-EtAl-2009-nlu-partial-sr},
determining when to respond during ongoing utterances \cite{devault-etal-2009-when-respond-inc-interpretation},
training actor policies for rapid task-based dialogue \cite{paetzel-etal-2015-inc-strategies},
continuous understanding and acting \cite{Stoness-2005-real-world-ref-for-slu},
incremental repair detection \cite{hough2014-inc-repair-detection},
or incremental reference resolution \cite{schlangenetal-2009-inc-ref-res}.

\section{Psycholinguistic evidence}
\label{sec:psych-evid}

Humans still pose the gold standard for language processing,
especially if the processing does not happen on large scale but in an
interactive setting.  When speaking, they perform several tasks
incrementally and in parallel, from conceptualization to articulation \cite{levelt-1989}.
Machines mimicking human behavior need to be
able to perform similar computations as humans in such settings to be
seen as a competent partner, and psycholinguistic research gives
insight into the processes that humans perform.
Psycholinguistic research is often carried out by timing experiments,
i.e. they make use of the incremental processing by humans to gain
insights into linguistic processes, mostly by eye-tracking
\cite{Tanenhaus-1995-int-vis-ling-info,sturtLombardo05,malsburg-2011-scanpath-syntactic-reanalysis}, but also by timing
word-by-word reading of sentences
\cite{gibson-2004-read-time-int-ling-struct}.

Language is perceived incrementally and this incremental processing is
influenced by other modalities even while speech is perceived
\cite{Tanenhaus-1995-int-vis-ling-info}.
Humans seem to create fully connected structures for sentence
prefixes, even for coordinate structures, and a continuation of a
sentence that does not match the predicted structure can be measured
by increased reading times \cite{sturtLombardo05}.  They perform
syntactic re-analysis and use various strategies to rescan sentences
\cite{malsburg-2011-scanpath-syntactic-reanalysis}.

\section{A typology for incremental problems}
\label{sec:types-data}

\begin{figure}
  \centering
  \def\akstretch{0.7ex}
  \begin{tabular}{rcl}
    \toprule
    Process & Data & \textbf{Data} / \emph{alignment} properties\\\midrule
            &\tikz[remember picture, baseline=(peterpa.base),->, >=latex]{
              \node (peterpa) {Peter};
              \node[base right= 0em of peterpa] (boughtpa) {bought};
              \node[base right= 0em of boughtpa] (flourpa) {[noun]};
              \coordinate[above = 3.5ex of boughtpa] (root);

              \draw  (boughtpa) to [in=60,out=120] (peterpa);
              \draw  (boughtpa) to [in=120,out=60] (flourpa);
              \draw  (root) to  (boughtpa);
              
              }    &  \textbf{structured, including prediction}\\[\akstretch]
    Parsing &      & \emph{vertices partially grounded, edges not grounded} \\[\akstretch]
            &
              \tikz[remember picture, baseline=(peteren.base)]{
              \node (peteren) {Peter};
              \node[base right= 0em of peteren] (bought) {bought};
              \node[base right= 0em of bought] (flour) {flour};
              }
                   & \textbf{discrete, sequential} \\[\akstretch]
    Machine translation & & \emph{reordering, fuzzy mapping} \\[\akstretch]
            &
              \tikz[remember picture, baseline=(peter.base)]{
              \node (peter){Peter};
              \node[base right= 0em of peter] (hat) {hat};
              \node[base right= 0em of hat] (mehl) {Mehl};
              \node[base right= 0em of mehl] (gekauft) {gekauft};
              }
                   & \textbf{discrete, sequential} \\[\akstretch]
    Speech recognition      & &\emph{order-preserving, clear mapping} \\[\akstretch]
            &
              \tikz[remember picture,baseline=(waveform.west)]{
                  \tikzstyle{curly} = [decorate,decoration={brace,amplitude=0.8ex},xshift=0em,yshift=0.5ex]

              \node (waveform) {\includegraphics[width=11em]{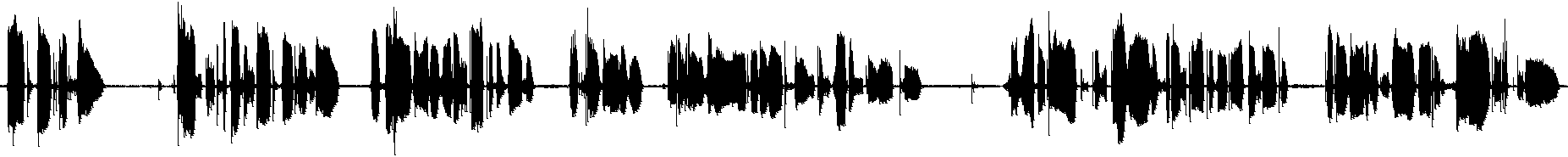}};
              \coordinate [below=1ex of waveform.north west] (wfwest);
              \coordinate [below=1ex of waveform.north east] (wfeast);

              \coordinate (wf-1) at ($(wfwest)!0.01!(wfeast)$);
              \coordinate (wf-2) at ($(wfwest)!0.23!(wfeast)$);
              \coordinate (wf-3) at ($(wfwest)!0.40!(wfeast)$);
              \coordinate (wf-4) at ($(wfwest)!0.70!(wfeast)$);
              \coordinate (wf-5) at ($(wfwest)!0.99!(wfeast)$);
              \foreach \wfanchor in {1,2,3,4} {
                \pgfmathtruncatemacro{\wfnextanchor}{\wfanchor+1}
                \draw [blueuhh,curly] (wf-\wfanchor) -- (wf-\wfnextanchor) node [midway,yshift=-0.5ex] (wf-brace-\wfanchor) {};
              }
              }
                   & \textbf{continuous}\\
    \bottomrule
  \end{tabular}
  \tikz[overlay, remember picture, style={color=blueuhh}]{
    \draw  (peter) to [in=-90,out=90] (peteren);
    \draw  (mehl) to [in=-90,out=90] (flour);
    \draw  (hat) to [in=-100,out=80] (bought);
    \draw  (gekauft) to [in=-80,out=100] (bought);

    \draw (peteren)  to [in=-90,out=90] (peterpa);
    \draw (bought)   to [in=-90,out=90] (boughtpa);

    \draw ($(wf-brace-1.north)+(0,0.5ex)$)  to [in=-90,out=90] (peter);
    \draw ($(wf-brace-2.north)+(0,0.5ex)$)  to [in=-90,out=90] (hat);
    \draw ($(wf-brace-3.north)+(0,0.5ex)$)  to [in=-90,out=90] (mehl);
    \draw ($(wf-brace-4.north)+(0,0.5ex)$)  to [in=-90,out=90] (gekauft);
  }
    \caption{Different types of data, groundings, and processors. Note
      that for parsing only the prefix ``Peter bought'' is processed
      to exemplify intermediate structure generated for incomplete input.
      Incremental systems can take many forms, and this example is not meant
      to perform a specific task.}
  \label{fig:datatypes}
\end{figure}

When working on an incremental processor, it is helpful to know which
other already existing processors had to deal with similar challenges
despite being designed for completely different NLP tasks.
In this section, I will examine properties relevant for classifying processors.
These properties describe the input and output data as well as the
relation between input and output
and help classifying
concrete tasks in Section~\ref{sec:incr-proc}.  For an in-depth
discussion of properties relevant to incremental processing, see Chapter 5 of
\newcite{guhe-2007-inc-conceptualization}\footnote{The properties
discussed by \newcite{guhe-2007-inc-conceptualization} are mostly complementary to the ones discussed here.}.

\subsection{Data types}
\label{sec:data-types}

Data can be \emph{structured} (e.g.\ syntactic or semantic structures) or
\emph{sequential}.  Sequential data can be \emph{discrete} (e.g.\ words) or
\emph{continuous} (e.g.\ speech signals) along the time axis\footnote{Data
  that is discrete along the time axis can of course include
  continuous data such as word embeddings.}. Structured data is always
discrete on the time axis. A processor can take one type of data as an input and create
another one as output; some examples are depicted in
Figure~\ref{fig:datatypes}.

Sequential data can usually be subdivided along a time axis,
i.e. there exists a total ordering between the elements of the input
(or output respectively).  Structured data does not exhibit this
property: Given, for example, the dependency tree produced by parsing in
Figure~\ref{fig:datatypes}, the words are ordered, but the
dependencies between the words can't be clearly attributed to a word:
They could be attributed to the head or the dependent; this poses an
additional challenge for evaluation.

\subsection{Granularity}
\label{sec:granularity}

The \emph{granularity} determines the size into which the input and output
is subdivided.  Assuming coarse enough granularity, every system can be
seen as incremental: A normal syntax parser works non-incremental inside a sentence,
but incrementally if processing a paragraph with
the basic units being sentences.  Grapheme to phoneme conversion is
incremental when processing text with the basic unit of words.  When considering a larger incremental
system, a processor usually seen as non-incremental might be incremental enough: If a
language generation system works on the level of words, the grapheme
to phoneme conversion does not need to be able to process sub-word
input.  On the other hand, if input typed by a user should be
vocalized, sub-word granularity might be needed.  The granularity of a
pipeline is determined by its most coarse-grained component.
In general, fine-grained processing is harder than coarse-grained
processing; a system can always process data fine-grained internally
while having coarse-grained interfaces, whereas the opposite is not
possible.

\subsection{Grounding}
\label{sec:grounding}

\emph{Grounding} describes the alignment from the generated
output to the elements of the input that yielded evidence for this
output \cite{schlangen-skantze-2009-incremental-units}.
Grounding allows to reason about which part of the output can be
reasonably generated given only partial input.
In some cases, this alignment is explicit in both test and
training data, e.g.\ in sequence labeling tasks where each element of
the input is assigned a label.  In other cases such as machine
translation, there is no gold standard word alignment\footnote{At least not
  on the word level, but alignments can be generated automatically, see e.g.\ \newcite{och03:asc}} and even a human-generated alignment would not
create a one-to-one mapping between input and output.  In addition,
the alignments may or may not be \emph{order-preserving}:
A tagging task preserves the ordering, whereas in translation reordering takes place
(cmp. ``hat Mehl gekauft'' \(\rightarrow\) ``bought flour'' in Figure~\ref{fig:datatypes}).

\subsection{Monotonicity}
\label{sec:monotonicity}

A system is \emph{non-monotonic} if it is allowed to retract output it
has previously produced.  For example, an incremental sequence labeler
that is free to re-assign labels can change its mind about every
element for a sentence once the sentence is complete.  A
\emph{monotonic system} on the other hand is required to only extend
previously generated output without retracting information.  Some
components are inherently monotonic because their output is not fed to
another processor of the system but to the outside world; e.g.\ a speech synthesizer is
inherently monotonic as it cannot retract sound waves realized through a loudspeaker.

Monotonicity limits the quality a component can produce as it can not
revert a decision that turns out to be wrong later on in light of
additional available input.  In contrast, a non-monotonic component
can always achieve the same non-incremental output as a
non-incremental component by simply replacing all intermediate output
with the one of the non-incremental component once all input is
available.

The precise meaning of monotonicity needs to be defined for each
component.  For sequential output, the most common definition is to
only allow appending to the output.  For structured output, the
structure of an increment could be required to be a super-set of its
predecessor.

Non-monotonic output can only be generated sensibly if the consumers
of the output can deal with non-monotonic input.  Otherwise these
consumers might ignore the revisions made to previous output and end up
with an inconsistent input or have to restart their computation in light of new input.

\subsection{Timeliness}
\label{sec:timeliness}

Each NLP processor has to optimize \emph{what} to output given an
input.  An incremental system also needs to decide \emph{when} to
provide output.  Discrete input provides specific anchors for this
decision, continuous input does not and new output can be generated
continuously\footnote{While continuous input is made discrete before
  reaching a processor, the discretization is usually in the order of
  milliseconds and can be seen as continuous for all practical
  purposes.}.
Such decisions also need to be made by human interpreters while
performing simultaneous translation; they need to buffer input until
they are able to produce additional output.  The characteristics of
this (human) process varies by language, e.g.\ the delay is relatively long
when translating from German to English because the verb in the input
tends to be delayed
\cite{Goldman-Eisler1972-segmentation-sim-translation}.

\subsection{Trade-Off between properties}
\label{sec:property-trade-off}

Incremental components have to make a trade-off between timeliness
(i.e.\ the amount of delay introduced between input and output),
output quality, and the amount of non-monotonicity
\cite{beuck11:_decis_strat_increm_pos_taggin,baumann-2009-speech-rec-inc-systems}.
High-quality, monotonic output can be obtained by delaying all output.
This strategy taken to the extreme results in a non-incremental system
that only produces one complete output once all input is available.
To reduce the delay, non-monotonicity via output revisions can be allowed, or
compromises with respect to the quality of the output can be made.
Gradual trade-offs can also be performed: allowing infrequent revisions and/or mild delays can
lessen the negative impact on accuracy.
These trade-offs are universal to all incremental processors.

\section{Incremental processors}
\label{sec:incr-proc}

This section discusses three tasks that can be performed incrementally
to exemplify strategies to ``incrementalize'' a processor:
first, incremental speech
recognition, a continuous sequence labeling problem that is usually
implemented as a non-monotonic processor, then  incremental machine
translation, which is a sequence to sequence task with reordering
implemented monotonically, and third parsing,
a sequence to structure task, which is implemented both monotonic an
non-monotonic.

\subsection{Speech recognition}
\label{sec:speech-recognition}

Speech recognition lends itself to incremental processing because it
is used in all interactive spoken dialogue systems and the decoding
happens incrementally even for non-incremental use-cases.  It is
therefore possible to look into a speech recognizer (SR) to obtain the
most probable hypothesis at each point in time without modifying the
recognizer \cite{baumann-2009-speech-rec-inc-systems}.  Because the SR
is not changed from the non-incremental one, the only optimization
point is when to let new output through, i.e.\ to implement a
\emph{gatekeeper}.  Its policy can be guided by observing the time
that a hypothesis survived without being discarded
\cite{baumann-2009-speech-rec-inc-systems} or based on the internal
state of the recognizer \cite{selfridge-EtAl:2011:stab-acc-inc-sr}.
\newcite{mcgraw-gruenstein-2012-estimating-word-stability-iasr} show
that even sophisticated stability estimation only slightly improves
upon the simple age-based estimation by \newcite{baumann-2009-speech-rec-inc-systems}.
Given that the SR is the same for the incremental case as for the
non-incremental one, no trade-off is performed against the accuracy.
In addition, \newcite{selfridge-EtAl:2011:stab-acc-inc-sr} noted that
if during decoding all beams in the beam search pass through the same state, the prefix up to that state can
be declared stable because future decoding will not change the most probable path
up to that state; this observation essentially makes use of the
Markov property and is primarily useful if grammar-based recognition is performed.

\subsection{Machine translation}
\label{sec:incr-mach-transl}

Translation can be performed as batch processing (e.g.\ for websites)
or incrementally, to facilitate human-human interaction.  Modern
approaches to machine translation, i.e.\ neural machine translation,
employ a sequence to sequence model where the input sequence is
encoded into a representation and then decoded again.  This can be
performed using recurrent networks which represent the input as a
single fixed-length vector and possibly modeling attention to the
input when generating the output sequence
\cite{Bahdanau-2014-nmt-joint-align-translate} or even without a
recurrent network, using only attention to model the influences from the
input sequence to the output sequence to generate
\cite{vaswani-2017-attention-all-you-need}.  In all these cases,
all input is consumed before translation happens.

\subsubsection{Incremental Neural Machine Translation (NMT) by using a gatekeeper}
\label{sec:incrementalizing-nmt}

\begin{figure}
  \centering
  \includegraphics[width=.5\textwidth]{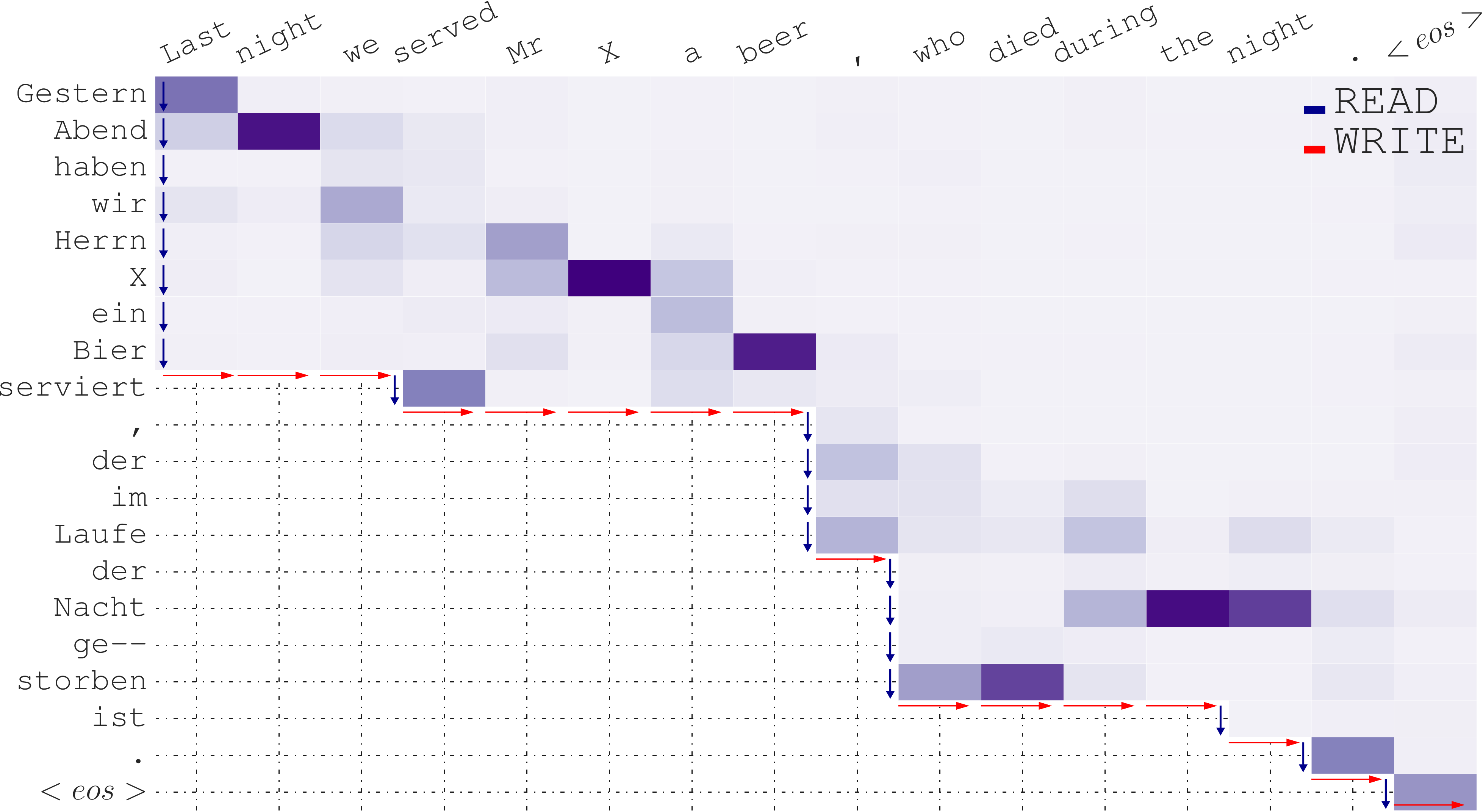}
  \hfill \raisebox{-1em}{\includegraphics[width=.40\textwidth]{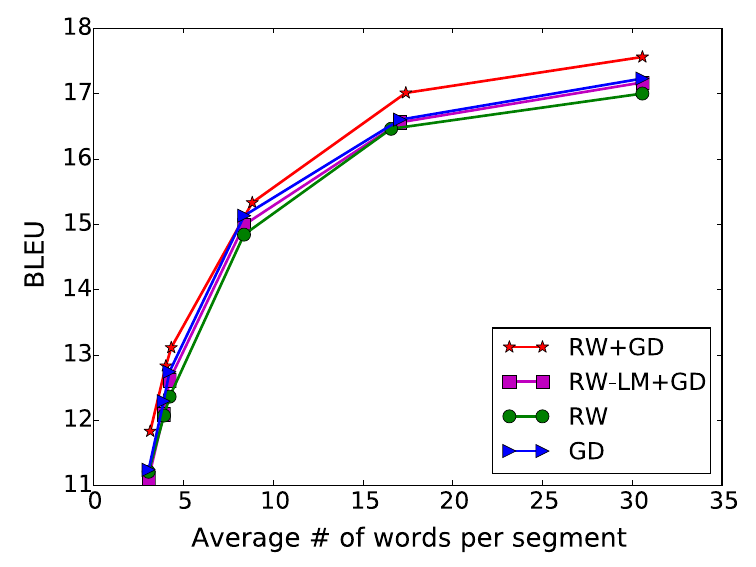}}
\caption{Left: Incremental NMT with attention, from
  \protect\newcite{gu-2017-real-time-nmt}.  Colors denote the
  attention given to each input token when generating an output
  token. Note the delay induced by the reordering for ``serviert''
  (\emph{served}) and ``gestorben'' (\emph{died}).
Right: trade-off decisions between delay (x-axis, measured in words translated at once) and accuracy (y-axis, measured in BLEU) from \protect\newcite{he-2015-syntax-rewriting-simult-mt}.}
  \label{fig:incremental-nmt-attention}
\end{figure}

As already discussed, machine translation is a sequence to sequence
problem where the output ordering does not conform to the input
ordering and the ground truth for the grounding often is missing.
As the incremental machine translation systems found in the
literature are monotonic, the systems need to decide at
which point they have enough information from the input to generate the
next output token with high confidence.
\newcite{gu-2017-real-time-nmt} propose a system where an NMT
processor repeatedly proposes an output token based on the currently
available input and the generated output to a gatekeeper.  The
gatekeeper either accepts this output, resulting in a write operation
to the output, or rejects it, resulting in a read operation on the
input.  The gatekeeper can work based on different policies, yielding
different trade-offs between timeliness and accuracy:
rejecting all output until the input is complete means
falling back to a non-incremental behavior; performing alternating
read and write actions eliminates delay but results in bad
translations.  \newcite{gu-2017-real-time-nmt} train the policy using
reinforcement learning with the reward based on the resulting BLEU
score and the delay incurred; weighting them differently results in
different trade-offs.  Humans have different preferences regarding
this trade-off, depending on whether the output is speech or subtitles \cite{mieno15interspeech}.
The underlying NMT model is trained on complete sentences and
therefore not adapted to incremental processing.  An example of an incremental
translation can be seen in Figure~\ref{fig:incremental-nmt-attention}.

\subsubsection{Reordering output}
\label{sec:reordering-output}
Reordering phenomena are a major hindrance to timely translation, as
exemplified in Figure~\ref{fig:incremental-nmt-attention}.
\newcite{GrissomII2014} deal with this by training a classifier to predict
verbs needed for the output but not yet seen in the input, allowing the MT
system to produce a verb without having seen its counterpart on the input.
\newcite{he-2015-syntax-rewriting-simult-mt} instead propose
to edit the gold
standard translations to better fit the source ordering for training
the MT system by trying to imitate the transformations a human translator would perform.
This approach tackles the problem that the training data
is only available for the non-incremental case, which is not optimal
for simultaneous translations.  Human simultaneous translators produce
sentences that systematically deviate from the ``normal'' target
language but such data is
not readily available for training \cite{he-etal-2016-human-strat-sim-interp}.
The training data is adapted by generating phrase-structure trees for the target
sentences and applying (manually written) syntax-based reordering
rules.  The system then checks whether the reordering has reduced the
delay based on an automatically computed alignment and if so uses the reordered version instead of the original
one.  As in the approach by \newcite{gu-2017-real-time-nmt}, the
average delay induced by the system can be tuned, see Figure~\ref{fig:incremental-nmt-attention}.
Because the processor 
was not trained on gold-standard data, it might yield sub-optimal
results on gold-standard test data.
\newcite{he-2015-syntax-rewriting-simult-mt} evaluate on both
gold-standard and transformed data and show that the system in
fact performs better on both targets.

\subsection{Parsing}
\label{sec:parsing}

Many state-of-the-art parsers work incrementally internally, both for
semantic and for syntactic parsing: they use a transition system
with a scorer and optionally combine that with beam search to find the best parse
\cite[\emph{inter alia}]{nivre-2008-deterministic-dep-parsing,huang-sagae:2010:ACL,dyer-EtAl:2015:ACL-IJCNLP,Swayamdipta2016,Kiperwasser:2016:bilstm-parsing,zhou-2016-amr-inc-joint-model,damonte-2017-inc-parser-amr}.
While they build a structure incrementally going from left to right,
they don't produce intermediate structures meant for incremental
consumption; the intermediate states consist of several unconnected
sub-structures.  In addition, they usually employ look-ahead which delays
the processing; approaches using Bi-LSTMs (such as \newcite{Kiperwasser:2016:bilstm-parsing})
effectively make use of the
whole sentence during each step, making the computation
non-incremental as it depends on the complete input being available.
\newcite{noji-2014-left-corner-dep-parsing} propose a transition
system based on the ones discussed by
\newcite{nivre-2008-deterministic-dep-parsing} that introduces dummy
nodes which denote an expectation of upcoming words.  This transition
system still produces disconnected trees for sentence prefixes but is
able to predict processing difficulties humans have when reading
sentences with center embeddings.  The other approaches described in
this section all produce connected structures for sentence prefixes.

\subsubsection{Incremental parsing with monotonic expansions}
\label{sec:incr-pars-with}

One approach to incremental parsing that produces connected structures at
each step is to monotonically extend a connected structure and to
employ a beam of possible structures to prevent being stuck with a
structure that does not fit the continuation of the sentence.
Using beam search, a parser does not provide monotonic
output but still guarantees that one of the beam entries
will be selected for a continuation.

\newcite{hassan-2009-semi-inc-ccg-parsing} show why either a beam
or delay is
necessary if performing incremental parsing with monotonic
extensions: They experiment with a parser based on Combinatory
Categorial Grammar \cite{Steedman-2000-syntactic-process}.  Their parser achieves an accuracy of
86\% when using lookahead and performing greedy parsing (i.e.\ it does
not use a beam).  This accuracy 
drops significantly
to 56\% without lookahead because the parser often commits to a structure
incompatible with the continuation of a sentence.

\newcite{Roark-2001-td-pars-lm} presents a top-down phrase
structure parser that performs beam-search to generate
connected intermediate structures for
every sentence prefix.  As the parser is based on a probabilistic
generative model, it can be used for language modeling and beats
trigram models on the Penn Treebank \cite{Marcus1994} (but not
other sequence models, see e.g.\ \newcite{merity-etal-2017-reg-opt-lstm-lms}).
\newcite{demberg-keller-2008-pltag} describe the
PLTAG formalism, which is based on Tree Adjoining Grammar
\cite{Joshi1997} and strives for psycholinguistic plausibility.  It
not only predicts upcoming structure needed for connectedness, but
also structure required by e.g.\ valencies not yet filled in the
prefix.  The predictions of the second type are automatically
extracted from the treebank during lexical induction by learning the
distinction between modifiers and arguments.
The PLTAG formalism provides better predictions for reading-time experiments
than predictions based on Roark’s parser \cite{demberg-2013-pltag}.
The combination of only extending entries in the beam and prediction leads to some sentences
being unparseable for both parsers because the structures stored in
the beam can all become incompatible with the input to be consumed.  A
larger beam reduces the number of unparseable sentences at the cost of
additional memory and processing cost, but can not eliminate the problem.

\subsubsection{Incremental gold standards for parsing}
\label{sec:incr-gold-stand}

\newcite{koehn14:incrTP} go in the opposite direction and perform
incremental parsing using restart-incrementality, i.e. performing a
complete new parse for each sentence prefix without reusing previously
generated output.  This approach gives no guarantees that a certain
structures will still be present in subsequent outputs but on the other
hand is able to react on any upcoming input without constraints, yielding
– in contrast to the approaches described in §~\ref{sec:incr-pars-with} –
the same accuracy for complete sentences as its non-incremental counterpart.  The
underlying parser performs a graph-based optimization
\cite{Martins2013} and has no notion of incrementality; instead, the
gold standard the parser is trained on is adapted to consist of
syntactic structures for sentence prefixes which contain prediction nodes
as stand-ins for upcoming words.
The partial dependency structure for a prefix is created by a rule
based system that works on the complete dependency structure for the
sentence.  It keeps all dependencies between words in the prefix and
delexicalizes words that are needed to connect the words in the prefix
to the sentence root as well as arguments that are deemed to be
predictable.  Words outside the prefix that are neither needed for
connection nor deemed to be predictable are deleted.
The downside of this
approach (in contrast to \newcite{demberg-keller-2008-pltag}, but similar to the reordering of output discussed in
§~\ref{sec:reordering-output}) is that a rule set needs to be
manually created for each language and that the linguistic intuition encoded into this
rule set might be a bottleneck for the parser as structures not
envisioned by the rule writer might be a better fit than the one
created by the rules.
\newcite{koehn-coling-2016} show that using the delexicalized
predictions to augment 5-gram language models improves perplexity on the
Billion Word Corpus \cite{chelba-etal-2013-bwc}.

\subsection{Natural language generation}
\label{sec:incr-conc}

\newcite{skantze-Hjalmarsson-2013-inc-speech-gen} compare a
non-incremental dialogue system and an incremental one, with which
language learners interact to buy items at a flea market.  The speech
recognition component was performed manually, i.e.\ a human manually
transcribed the speech.
As the manual transcription takes time, the system response is
noticeably delayed in a non-incremental system where the dialogue system
only starts to plan its response once transcription is complete.
In contrast, the incremental system constructs a response as soon as
possible, based on partial input.  The response is recomputed
on changed input, which can have three effects: If
the update is consistent with what has been said already, the
continuation of what to say is simply changed (a \emph{covert
  change}). If the system has to retract information already uttered,
an explicit repair has to be produced (an \emph{overt change}).  If
not all information to produce a complete utterance is available,
fillers are inserted to avoid silence.  The system is faster and preferred by users
even though it has to explicitly correct itself, resulting in longer
responses.  Even though its output is monotonic – as it cannot erase
information from the hearer’s ears – its explicit repairs allow to act
with low delay (a similar strategy to the one employed by human
speakers \cite[ch. 12]{levelt-1989}).  As the NLG component is rule-based,
it is not constrained by the (non-) availability of data suitable for
learning incremental language generation.

\section{Evaluating incremental systems}
\label{sec:eval-incr-syst}

An incremental system can be evaluated just as a non-incremental one.
Evaluation schemata exist for all established tasks such as speech
recognition (quality measured in word error rate), PoS tagging
(measured in accuracy), phrase structure parsing (measured in
precision/recall), or machine translation (BLEU).  Additional
evaluation allows to examine the behavior more closely, such as
compiling error confusion matrices.  These schemata enable a
comparison between components in a standardized way.

The downside of standardized evaluation is the lack of insight for
incremental properties.  When building incremental systems, not only the final output but also
the intermediate behavior is of importance but using
evaluations tailored to non-incremental systems fails to give insight into the incremental properties of
the system.
An incremental system should provide as much correct information as
early as possible but using non-incremental evaluation hides all
differences in this respect.

\definecolor{hlcolor}{rgb}{0.9,0.9,0.9}
\definecolor{hlcolor2}{rgb}{0.35,0.35,0.35}
\newcommand{\spc}{\hspace{0.3em} }
\begin{table}
  \centering
  \begin{tabular}{rllll}
    \toprule
         & non-monotonic (1) & non-monotonic (2) & delayed (3) & erroneous (4)\\
\ncoord[0,1.6ex]{timestart}\hfill
input: a & \ncoord[-0.2em,0ex]{a1}a/\incorrect{y}\ncoord[-0.2ex,0ex]{a1e}\spc  & a/\incorrect{y}\spc  &             & a/\incorrect{y}\spc        \\
       b & a/\correct{x}\spc b/\correct{y}\spc          & a/\incorrect{y}\spc b/\correct{y}\spc & a/\correct{x}\spc b/\correct{y}\spc    & a/\incorrect{y}\spc b/\correct{y}\spc    \\
\ncoord{timestop}\hfill
       c & \ncoord[0,1.2ex]{a3}a/\correct{x}\spc b/\correct{y}\spc c/\correct{z}\spc      & a/\correct{x}\spc b/\correct{y}\spc c/\correct{z}\spc      & a/\correct{x}\spc b/\correct{y}\spc c/\correct{z}\spc & a/\incorrect{y}\spc b/\correct{y}\spc c/\correct{z} \\\midrule
inc\_acc(i) & 2: \nicefrac{1}{1};\ 1: \nicefrac{2}{2};\ 0: \ncoord[0.3ex,0]{i0}\nicefrac{2}{3}\ncoord[0.5em,0ex]{i0e} & 2: \nicefrac{1}{1};\ 1: \nicefrac{1}{2} ;\ 0: \nicefrac{2}{3} & 2: \nicefrac{1}{1};\ 1: \nicefrac{2}{2} ;\ 0: \nicefrac{2}{3} & 2: \nicefrac{0}{1};\ 1: \nicefrac{1}{2} ;\ 0: \nicefrac{2}{3}  \\
EO       & \nicefrac{1}{4}   & \nicefrac{1}{2}   & 0           & 0 \\
accuracy & \nicefrac{3}{3}  & \nicefrac{3}{3}    & \nicefrac{3}{3} & \nicefrac{2}{3}  \\
    \bottomrule
  \end{tabular}
  \tikz[overlay,remember picture, blend mode = soft light] {
    \draw[->] (timestart) -- (timestop);
    \draw[rounded corners=0.3em, dashed, thick, draw, greyuhh]
    ($ (a1) + (0,1.8ex) $) --
    ($ (a1e) + (0,1.8ex) $) --
    ($ (i0e) + (0,1.6ex) $) --
    ($ (i0e) + (0,-0.4ex) $) --
    ($ (i0) + (0,-0.4ex) $) --
    ($ (a1) + (0,0.0ex) $) --
    cycle;
    node[above, midway, rotate=90] {\small{time}};
  }
  \caption[example characteristics for incremental output]{Examples
    for characteristics of incremental output that need to be
    captured for evaluation.
    Correct output: a/x b/y c/z.
    (1), (2): output for \emph{a} changed;
    (3): output for \emph{a} held back until input ``b'' is available;
    (4): incorrect assignment to ``a'' stays in output.
    Dashed: exemplifies output used to compute inc\_acc (incremental accuracy).
    EO: edit overhead. Accuracy: measured on complete output.}
  \label{tab:seq-example-characteristics}
\end{table}

Table~\ref{tab:seq-example-characteristics} shows abstract
input/output patterns for a sequence labeling task, where the
correspondence between input and output is given and no reordering
effects take place.
A standard
accuracy-based evaluation on the complete output would yield a perfect
score for the three first systems although their behavior over time is
quite different:
In addition to the non-incremental evaluation for the
complete output, the non-monotonicity in (1) and (2) as well as the
delay in (3) need to be covered.
It is impossible to boil down all these differences to a single number.
We will therefore have a look at distinct metrics for timeliness,
monotonicity, and  quality.

\subsection{Measuring timeliness}
\label{sec:measuring-timeliness}

\newcite{cho-esipova-2016-simultaneous-nmt} propose to measure timeliness
by counting for each output element \(t\) of an output sequence \(Y\)
how many input elements from the input sequence \(X\) have been
consumed before its production (\(s(t)\)).  \(\tau(X,Y)\) then
computes the translation delay:

\[0 < \tau(X,Y) = \frac{1}{|X||Y|}\sum_{t=1}^{|Y|}s(t) \leq 1 \]

This computation is helpful when there is no gold standard alignment
between output and input which one could use to obtain the output timing
that could be achieved under optimal conditions.
\(\tau=0\) means all output was made without consuming input,
\(\tau=1\) means all input was read before generating output.

\newcite{GrissomII2014} introduce latency-BLEU, a metric that averages
the BLEU scores of the outputs corresponding to each input prefix.
The complete translation is weighed higher than all other partial
translations to penalize incorrect translation.  If the resulting
translation is the same, a processor with less delay will obtain a
higher score.  It has to be noted that due to the averaging the
sentence-initial output has more influence on the score than output
created near the end of a sequence.  It is also not possible to
completely distinguish between the quality and the timeliness because
quality and timeliness are measured in a single metric.
In the MT system by \newcite{he-2015-syntax-rewriting-simult-mt}, the
timeliness can be (indirectly) tuned by adjusting a threshold at which
all yet untranslated words should be translated.
Figure~\ref{fig:incremental-nmt-attention} (right) shows a plot of the
resulting BLEU score against the average number of words translated at
once, i.e. the delay.  This way, potential users can see the
trade-offs that can be made using a system  (RW+GD is the proposed
architecture, beating the other approaches at each trade-off point).

If an explicit alignment exists between the input and the output –
such as in speech recognition – the difference between when an output
is made (i.e. which amount of input data has been consumed) and the
timing of the corresponding input can be measured to obtain an
anchored timeliness measure \cite{baumannetal2011:dnd}.  Both the
relation to the first occurrence of an output (\texttt{FO}) and the
relation to the last change of an output (final decision, \texttt{FD})
can be measured.
E.g.\ if a word ends at 2.5 seconds of the input
audio, was first recognized after consuming 3 seconds and was part of
all outputs produced after consuming 4 seconds, its
\texttt{FO} would be 0.5 seconds and its \texttt{FD} would be 1.5 seconds.
To separate the timeliness from the quality, these measures
can be computed against the final output of the system instead of the
gold standard,  but then a
reliable alignment between the input and the generated output is
needed.  The advantage to other methods is its interpretability: A
\texttt{FO} of 100ms for a speech recognizer means that it produces,
on average, an output 100ms after it has consumed
input that carries evidence for this output.  Obviously, \texttt{FO} and
\texttt{FD} only differ for non-monotonic processors; \texttt{FD}
measures what delay is necessary on average to rely on an output
produced by the processor.

\subsection{Measuring incremental quality}
\label{sec:meas-incr-accur}

When dealing with monotonic output, the incremental quality
can be assessed using the non-incremental quality metrics as the processor can't correct
previously made output.
If a processor can be tuned to provide more or less timely output,
the quality can be plotted against delay, as in Figure~\ref{fig:incremental-nmt-attention}.

If a system is non-monotonic, it makes sense to measure the accuracy
not only based on the final output but also on the intermediate output.
Averaging the quality for each increment has two downsides: Output for
early input is weighed more heavily than for later input, and it is unclear
how the non-monotonicity affects the quality.
\newcite{beuck:11:pda} perform an evaluation for incremental
dependency parsing by computing the accuracy for the \emph{n}-th word
to the right of the frontier in every prefix, with n between zero
(measuring the newest words) and five (the accuracy of the
sixth-newest word).
This way, an accuracy curve relative to the age of the input is generated,
Table~\ref{tab:seq-example-characteristics} shows incremental accuracy (\emph{inc\_acc})
measures relative to the age of the input; the examples (1) and (2)
obtain different incremental accuracy measures even though the
non-incremental accuracy is the same.

Incremental structured output might include predictions, which are
not accounted for in the metrics discussed
up to now.  If an incremental gold standard is available (such as the
one used for training discussed in §~\ref{sec:incr-gold-stand}), the
precision and recall of those predictions can be computed with respect
to the predictions in the gold standard \cite{beuck13:PredIncr}.

\subsection{Measuring the degree of non-monotonicity}
\label{sec:measuring-amount-non}

Evaluating non-monotonicity can be viewed from two (similar)
standpoints: first, how much intermediate output will be retracted
again? And second: how sure can we be that a certain output is
reliable, i.e.\ will also be part of the final output of a processor?

\newcite{baumann-2009-speech-rec-inc-systems} and  \newcite{Baumann2013} tackle the first question by defining the
\emph{edit overhead} generated by a non-monotonic processor producing
sequential output by defining three edit operations on an output
sequence: \emph{add} (append an element to the output), \emph{revoke}
(remove the last element from the output), and \emph{substitute}
(revoke and then append). The difference \(\textit{diff}(o_{i},o_{j})\) between
two outputs \(o_{i}\) and \(o_{j}\) can then be defined as the minimal
number of edits needed to change \(o_{i}\) into \(o_{j}\).  Note that to
change the first element of an output of length \(n\), \(2(n-1)+1\)
operations are needed whereas changing the last element only yields a
difference of one.  This notation allows to compute the edit overhead:
Let \(N_{optimal}=|\textit{diff}(o_{0}, o_{tmax})|\) be the number of edits
necessary to obtain the final output and
\(N_{actual}=\sum_{t=1}^{tmax}\textit{diff}(o_{t-1},o_{t})\) be the amount of edit operations
actually performed.  The edit overhead is then defined as the
proportion of unnecessary edits produced by the processor:
\[EO =
  (N_{actual}-N_{optimal})/N_{actual}\]

Table~\ref{tab:seq-example-characteristics} shows that \(EO\) can
distinguish between the different levels of non-monotonicity.
The edit operations can be adapted to the problem at hand, e.g.\ if
structured output is produced or edits at the start of the sequence
should not be penalized heavier than edits at the end.

Regarding the second question, \newcite{beuck:11:pda} propose to
compute the accuracy against the output generated based on the
complete input instead of computing against the gold standard to
obtain a stability measure.  Using the gold standard as reference
measures consistency with the ground truth, i.e. the quality, and
using the complete output measures the consistency with that, i.e. the
stability.
This approach obviously only works if an incremental accuracy is
defined for the problem at hand and that accuracy can be measured with
respect to a non-incremental gold standard.

\section{Combining multiple incremental processors}
\label{sec:comb-mult-incr}

A NLP system usually consist of several processors.
In a non-incremental system, all processors can simply form a pipeline with
each processor working on the output of the previous one.  This is
non-trivial in an incremental system because for a given input, each
processor may produce several outputs which may even be contradictory
due to non-monotonicity.  Therefore,
a system either needs to use restart-incrementality throughout the
pipeline or track changes to perform partial recomputation.
\newcite[ch. 5]{wiren-1992-thesis}
introduces the notion of dependencies to track which parts of the
input a certain output is based on in a chart parser.  This way, only
parts of the chart need to be recomputed given a non-monotonic change.
\newcite{schlangen-skantze-2009-incremental-units} vastly extend this
notion to a general computation model in which multiple processors
work on data which is organized in incremental units (IU).  An IU
stands for a minimal unit of information, such as a recognized word.
IUs are grounded in other IUs from a previous level, and linked to
other IUs on the same level, e.g.\ to describe a sequential
relationship.  Processors create new IUs based on their input and may
revoke IUs already generated.  A processor can commit to an IU to
signify that this IU will not be revoked and other processors may rely
on it.

\section{Conclusion and outlook}
\label{sec:conclusion}

Building incremental processors poses problems that are similar
regardless of the domain.  Decisions have to be made regarding the amount
of delay acceptable, whether non-monotonic output is acceptable for downstream processors,
and if so, to what extent.
For all these decisions, the relation between input and output is
important: Is there a clear mapping between input and output, maybe
even a one-to-one mapping, and does reordering happen?
Several techniques have been discussed for implementing incremental processors:
Training a \emph{gatekeeper} to either perform
a trade-off between non-monotonicity and delay
(§~\ref{sec:speech-recognition}) or – for monotonic processors – between
delay and quality (§~\ref{sec:incrementalizing-nmt}),
\emph{Beam search} to create non-monotonic output with monotonic extension (§~\ref{sec:incr-pars-with}),
and \emph{restart incrementality} for unrestricted non-monotonicity (§~\ref{sec:incr-gold-stand}).
To evaluate an incremental system, ideally three characteristics
should be measured: timeliness (§~\ref{sec:measuring-timeliness}), quality (§~\ref{sec:meas-incr-accur}), and non-monotonicity (§~\ref{sec:measuring-amount-non}).
If a system can be tuned in regard to these characteristics, different trade-off points
between these properties can be measured (§~\ref{sec:property-trade-off}).

There are still open problems for building incremental NLP systems:
Processors generating connected structured output need incremental
training data to learn intermediate structures not visible in the
non-incremental gold standards; as the quality of the training data
generation influences the quality of the processor, data-driven approaches
producing high-quality incremental training data are needed.
Explicit corrections in spoken
output are preferred by users to delay (§~\ref{sec:incr-conc}), which could transfer to
incremental MT.  However, lack of automatic evaluation likely requires an
expensive human end-to-end evaluation due to the large deviation from non-incremental
gold standard translations.
All data-driven processors discussed are able to produce non-monotonic output,
but are not able to consume non-monotonic input, a
problematic discrepancy for building incremental systems out of
multiple processors.
To bridge the gap between certainty (monotonic output) and uncertainty
(non-monotonic output), the likelihood of an output being stable could
be attached to the output (see \newcite{selfridge-EtAl:2011:stab-acc-inc-sr} §~5).
Modern NLP processors heavily rely on sub-symbolic representation and
use attention mechanisms to obtain the relevant information.  With
this approach, an explicit grounding for partial recomputation as discussed in
Section~\ref{sec:comb-mult-incr} does not work anymore and would need
to be replaced with some notion of soft grounding.

\section*{Acknowledgments}
\label{sec:acknowledgments}
I would like to thank Christine Köhn, Sebastian Beschke, and Timo
Baumann for valuable feedback, as well as the anonymous reviewers for
helpful remarks.

\bibliographystyle{acl}
\bibliography{incremental-nlp}

\begin{thebibliography}{}

\bibitem[\protect\citename{Aït-Mokhtar \bgroup et al.\egroup
  }2002]{ait-mokhtar-2002-robustnes-inc-deep-parsing}
S.~Aït-Mokhtar, J.-P. Chanod, and C.~Roux.
\newblock 2002.
\newblock Robustness beyond shallowness: incremental deep parsing.
\newblock {\em Natural Language Engineering}, 8(2-3):121–144.

\bibitem[\protect\citename{Bahdanau \bgroup et al.\egroup
  }2014]{Bahdanau-2014-nmt-joint-align-translate}
Dzmitry Bahdanau, Kyunghyun Cho, and Yoshua Bengio.
\newblock 2014.
\newblock Neural machine translation by jointly learning to align and
  translate.
\newblock {\em CoRR}, abs/1409.0473.

\bibitem[\protect\citename{Baumann \bgroup et al.\egroup
  }2009]{baumann-2009-speech-rec-inc-systems}
Timo Baumann, Michaela Atterer, and David Schlangen.
\newblock 2009.
\newblock Assessing and improving the performance of speech recognition for
  incremental systems.
\newblock In {\em Proceedings of Human Language Technologies: The 2009 Annual
  Conference of the North American Chapter of the Association for Computational
  Linguistics}, pages 380--388, Boulder, Colorado, June. Association for
  Computational Linguistics.

\bibitem[\protect\citename{Baumann \bgroup et al.\egroup
  }2011]{baumannetal2011:dnd}
Timo Baumann, Okko Buß, and David Schlangen.
\newblock 2011.
\newblock Evaluation and optimisation of incremental processors.
\newblock {\em Dialogue \& Discourse}, 2(1):113--141.
\newblock Special Issue on Incremental Processing in Dialogue.

\bibitem[\protect\citename{Baumann}2013]{Baumann2013}
Timo Baumann.
\newblock 2013.
\newblock {\em Incremental Spoken Dialogue Processing: Architecture and
  Lower-level Components}.
\newblock {Ph.D.} thesis, Universität Bielefeld, Germany.

\bibitem[\protect\citename{Beuck \bgroup et al.\egroup
  }2011a]{beuck11:_decis_strat_increm_pos_taggin}
Niels Beuck, Arne Köhn, and Wolfgang Menzel.
\newblock 2011a.
\newblock Decision strategies for incremental pos tagging.
\newblock In Bolette~Sandford Pedersen, Gunta~Ne\v spore, and Inguna Skadina,
  editors, {\em Proceedings of the 18th Nordic Conference of Computational
  Linguistics NODALIDA 2011}, volume~11 of {\em NEALT Proceedings}, pages
  26--33. Northern European Association for Language Technology (NEALT).

\bibitem[\protect\citename{Beuck \bgroup et al.\egroup }2011b]{beuck:11:pda}
Niels Beuck, Arne Köhn, and Wolfgang Menzel.
\newblock 2011b.
\newblock Incremental parsing and the evaluation of partial dependency
  analyses.
\newblock In {\em Proceedings of the 1st International Conference on Dependency
  Linguistics}. Depling 2011.

\bibitem[\protect\citename{Beuck \bgroup et al.\egroup }2013]{beuck13:PredIncr}
Niels Beuck, Arne Köhn, and Wolfgang Menzel.
\newblock 2013.
\newblock Predictive incremental parsing and its evaluation.
\newblock In Kim Gerdes, Eva Hajičová, and Leo Wanner, editors, {\em
  Computational Dependency Theory}, volume 258 of {\em Frontiers in Artificial
  Intelligence and Applications}, pages 186 -- 206. IOS press.

\bibitem[\protect\citename{Chelba \bgroup et al.\egroup
  }2013]{chelba-etal-2013-bwc}
Ciprian Chelba, Tomas Mikolov, Mike Schuster, Qi~Ge, Thorsten Brants, and
  Phillipp Koehn.
\newblock 2013.
\newblock One billion word benchmark for measuring progress in statistical
  language modeling.
\newblock {\em CoRR}, abs/1312.3005.

\bibitem[\protect\citename{Cho and
  Esipova}2016]{cho-esipova-2016-simultaneous-nmt}
Kyunghyun Cho and Masha Esipova.
\newblock 2016.
\newblock Can neural machine translation do simultaneous translation?
\newblock {\em CoRR}, abs/1606.02012.

\bibitem[\protect\citename{Damonte \bgroup et al.\egroup
  }2017]{damonte-2017-inc-parser-amr}
Marco Damonte, Shay~B. Cohen, and Giorgio Satta.
\newblock 2017.
\newblock An incremental parser for abstract meaning representation.
\newblock In {\em Proceedings of the 15th Conference of the European Chapter of
  the Association for Computational Linguistics: Volume 1, Long Papers}, pages
  536--546, Valencia, Spain, April. Association for Computational Linguistics.

\bibitem[\protect\citename{Demberg and Keller}2008]{demberg-keller-2008-pltag}
Vera Demberg and Frank Keller.
\newblock 2008.
\newblock A psycholinguistically motivated version of tag.
\newblock In {\em Proceedings of the Ninth International Workshop on Tree
  Adjoining Grammar and Related Frameworks (TAG+9)}, pages 25--32, Tübingen,
  Germany, June.

\bibitem[\protect\citename{Demberg \bgroup et al.\egroup
  }2013]{demberg-2013-pltag}
Vera Demberg, Frank Keller, and Alexander Koller.
\newblock 2013.
\newblock Incremental, predictive parsing with psycholinguistically motivated
  tree-adjoining grammar.
\newblock {\em Computational Linguistics}, 39(4):1025--1066.

\bibitem[\protect\citename{DeVault \bgroup et al.\egroup
  }2009]{devault-etal-2009-when-respond-inc-interpretation}
David DeVault, Kenji Sagae, and David Traum.
\newblock 2009.
\newblock Can i finish? learning when to respond to incremental interpretation
  results in interactive dialogue.
\newblock In {\em Proceedings of the SIGDIAL 2009 Conference}, pages 11--20,
  London, UK, September. Association for Computational Linguistics.

\bibitem[\protect\citename{Dyer \bgroup et al.\egroup
  }2015]{dyer-EtAl:2015:ACL-IJCNLP}
Chris Dyer, Miguel Ballesteros, Wang Ling, Austin Matthews, and Noah~A. Smith.
\newblock 2015.
\newblock Transition-based dependency parsing with stack long short-term
  memory.
\newblock In {\em Proceedings of the 53rd Annual Meeting of the Association for
  Computational Linguistics and the 7th International Joint Conference on
  Natural Language Processing (Volume 1: Long Papers)}, pages 334--343,
  Beijing, China, July. Association for Computational Linguistics.

\bibitem[\protect\citename{Gibson and
  Warren}2004]{gibson-2004-read-time-int-ling-struct}
Edward Gibson and Tessa Warren.
\newblock 2004.
\newblock Reading-time evidence for intermediate linguistic structure in
  long-distance dependencies.
\newblock {\em Syntax}, 7(1):55--78.

\bibitem[\protect\citename{Goldman-Eisler}1972]{Goldman-Eisler1972-segmentation-sim-translation}
Frieda Goldman-Eisler.
\newblock 1972.
\newblock Segmentation of input in simultaneous translation.
\newblock {\em Journal of Psycholinguistic Research}, 1(2):127--140, Jun.

\bibitem[\protect\citename{Grissom~II \bgroup et al.\egroup
  }2014]{GrissomII2014}
Alvin Grissom~II, He~He, Jordan Boyd-Graber, John Morgan, and Hal
  Daum\'{e}~III.
\newblock 2014.
\newblock Don\'t until the final verb wait: Reinforcement learning for
  simultaneous machine translation.
\newblock In {\em Proceedings of the 2014 Conference on Empirical Methods in
  Natural Language Processing (EMNLP)}, pages 1342--1352, Doha, Qatar, October.
  Association for Computational Linguistics.

\bibitem[\protect\citename{Gu \bgroup et al.\egroup
  }2017]{gu-2017-real-time-nmt}
Jiatao Gu, Graham Neubig, Kyunghyun Cho, and Victor~O.K. Li.
\newblock 2017.
\newblock Learning to translate in real-time with neural machine translation.
\newblock In {\em Proceedings of the 15th Conference of the European Chapter of
  the Association for Computational Linguistics: Volume 1, Long Papers}, pages
  1053--1062, Valencia, Spain, April. Association for Computational
  Linguistics.

\bibitem[\protect\citename{Guhe}2007]{guhe-2007-inc-conceptualization}
Markus Guhe.
\newblock 2007.
\newblock {\em Incremental Conceptualization for Language Production}.
\newblock Lawrence Erlbaum Associates, Inc.

\bibitem[\protect\citename{Hassan \bgroup et al.\egroup
  }2009]{hassan-2009-semi-inc-ccg-parsing}
Hany Hassan, Khalil Sima'an, and Andy Way.
\newblock 2009.
\newblock Lexicalized semi-incremental dependency parsing.
\newblock In {\em Proceedings of the International Conference RANLP-2009},
  pages 128--134, Borovets, Bulgaria, September. Association for Computational
  Linguistics.

\bibitem[\protect\citename{He \bgroup et al.\egroup
  }2015]{he-2015-syntax-rewriting-simult-mt}
He~He, Alvin Grissom~II, John Morgan, Jordan Boyd-Graber, and Hal
  Daum\'{e}~III.
\newblock 2015.
\newblock Syntax-based rewriting for simultaneous machine translation.
\newblock In {\em Proceedings of the 2015 Conference on Empirical Methods in
  Natural Language Processing}, pages 55--64, Lisbon, Portugal, September.
  Association for Computational Linguistics.

\bibitem[\protect\citename{He \bgroup et al.\egroup
  }2016]{he-etal-2016-human-strat-sim-interp}
He~He, Jordan Boyd-Graber, and Hal Daum\'{e}~III.
\newblock 2016.
\newblock Interpretese vs. translationese: The uniqueness of human strategies
  in simultaneous interpretation.
\newblock In {\em Proceedings of the 2016 Conference of the North American
  Chapter of the Association for Computational Linguistics: Human Language
  Technologies}, pages 971--976, San Diego, California, June. Association for
  Computational Linguistics.

\bibitem[\protect\citename{Hough and
  Purver}2014]{hough2014-inc-repair-detection}
Julian Hough and Matthew Purver.
\newblock 2014.
\newblock Strongly incremental repair detection.
\newblock In {\em Proceedings of the 2014 Conference on Empirical Methods in
  Natural Language Processing (EMNLP)}, pages 78--89, Doha, Qatar, October.
  Association for Computational Linguistics.

\bibitem[\protect\citename{Huang and Sagae}2010]{huang-sagae:2010:ACL}
Liang Huang and Kenji Sagae.
\newblock 2010.
\newblock Dynamic programming for linear-time incremental parsing.
\newblock In {\em Proceedings of the 48th Annual Meeting of the Association for
  Computational Linguistics}, pages 1077--1086, Uppsala, Sweden.

\bibitem[\protect\citename{Joshi and Schabes}1997]{Joshi1997}
Aravind~K. Joshi and Yves Schabes.
\newblock 1997.
\newblock Tree-adjoining grammars.
\newblock In Grzegorz Rozenberg and Arto Salomaa, editors, {\em Handbook of
  Formal Languages: Volume 3 Beyond Words}, pages 69--123. Springer Berlin
  Heidelberg, Berlin, Heidelberg.

\bibitem[\protect\citename{Kay \bgroup et al.\egroup }1994]{kay-1994-verbmobil}
Martin Kay, Jean~Mark Gawron, and Peter Norvig.
\newblock 1994.
\newblock {\em Verbmobil: a translation system for face-to-face dialog}.
\newblock Number~33 in CSLI lecture notes. CSLI.

\bibitem[\protect\citename{Kiperwasser and
  Goldberg}2016]{Kiperwasser:2016:bilstm-parsing}
Eliyahu Kiperwasser and Yoav Goldberg.
\newblock 2016.
\newblock Simple and accurate dependency parsing using bidirectional lstm
  feature representations.
\newblock {\em Transactions of the Association for Computational Linguistics},
  4:313--327.

\bibitem[\protect\citename{K{\"o}hn and Baumann}2016]{koehn-coling-2016}
Arne K{\"o}hn and Timo Baumann.
\newblock 2016.
\newblock Predictive incremental parsing helps language modeling.
\newblock In {\em Proceedings of COLING 2016, the 26th International Conference
  on Computational Linguistics: Technical Papers}, pages 268--277. The COLING
  2016 Organizing Committee.

\bibitem[\protect\citename{K\"{o}hn and Menzel}2014]{koehn14:incrTP}
Arne K\"{o}hn and Wolfgang Menzel.
\newblock 2014.
\newblock Incremental predictive parsing with turboparser.
\newblock In {\em Proceedings of the 52nd Annual Meeting of the Association for
  Computational Linguistics (Volume 2: Short Papers)}, pages 803--808,
  Baltimore, Maryland, June. Association for Computational Linguistics.

\bibitem[\protect\citename{Levelt}1989]{levelt-1989}
Wilem J.~M. Levelt.
\newblock 1989.
\newblock {\em Speaking: From Intention to Articulation}.
\newblock The MIT Press.

\bibitem[\protect\citename{Marcus \bgroup et al.\egroup }1994]{Marcus1994}
Mitchell Marcus, Grace Kim, Mary~Ann Marcinkiewicz, Robert MacIntyre, Ann Bies,
  Mark Ferguson, Karen Katz, and Britta Schasberger.
\newblock 1994.
\newblock {The Penn Treebank}: Annotating predicate argument structure.
\newblock In {\em Proceedings of the Workshop on Human Language Technology},
  HLT '94, pages 114--119, Stroudsburg, PA, USA. Association for Computational
  Linguistics.

\bibitem[\protect\citename{Martins \bgroup et al.\egroup }2013]{Martins2013}
Andr{\'e} Martins, Miguel Almeida, and Noah~A. Smith.
\newblock 2013.
\newblock Turning on the turbo: Fast third-order non-projective turbo parsers.
\newblock In {\em Proceedings of the 51st Annual Meeting of the Association for
  Computational Linguistics (Volume 2: Short Papers)}, pages 617--622, Sofia,
  Bulgaria, August.

\bibitem[\protect\citename{McGraw and
  Gruenstein}2012]{mcgraw-gruenstein-2012-estimating-word-stability-iasr}
Ian McGraw and Alexander Gruenstein.
\newblock 2012.
\newblock Estimating word-stability during incremental speech recognition.
\newblock In {\em Interspeech}.

\bibitem[\protect\citename{Merity \bgroup et al.\egroup
  }2017]{merity-etal-2017-reg-opt-lstm-lms}
Stephen Merity, Nitish~Shirish Keskar, and Richard Socher.
\newblock 2017.
\newblock Regularizing and optimizing {LSTM} language models.
\newblock {\em CoRR}, abs/1708.02182.

\bibitem[\protect\citename{Mieno \bgroup et al.\egroup
  }2015]{mieno15interspeech}
Takashi Mieno, Graham Neubig, Sakriani Sakti, Tomoki Toda, and Satoshi
  Nakamura.
\newblock 2015.
\newblock Speed or accuracy? a study in evaluation of simultaneous speech
  translation.
\newblock In {\em 16th Annual Conference of the International Speech
  Communication Association (InterSpeech 2015)}, Dresden, Germany, September.

\bibitem[\protect\citename{Nivre}2008]{nivre-2008-deterministic-dep-parsing}
Joakim Nivre.
\newblock 2008.
\newblock Algorithms for deterministic incremental dependency parsing.
\newblock {\em Computational Linguistics}, 34(4):513--553.

\bibitem[\protect\citename{Noji and
  Miyao}2014]{noji-2014-left-corner-dep-parsing}
Hiroshi Noji and Yusuke Miyao.
\newblock 2014.
\newblock Left-corner transitions on dependency parsing.
\newblock In {\em Proceedings of COLING 2014, the 25th International Conference
  on Computational Linguistics: Technical Papers}, pages 2140--2150, Dublin,
  Ireland, August. Dublin City University and Association for Computational
  Linguistics.

\bibitem[\protect\citename{Och and Ney}2003]{och03:asc}
Franz~Josef Och and Hermann Ney.
\newblock 2003.
\newblock A systematic comparison of various statistical alignment models.
\newblock {\em Computational Linguistics}, 29(1):19--51.

\bibitem[\protect\citename{Paetzel \bgroup et al.\egroup
  }2015]{paetzel-etal-2015-inc-strategies}
Maike Paetzel, Ramesh Manuvinakurike, and David DeVault.
\newblock 2015.
\newblock "so, which one is it?" the effect of alternative incremental
  architectures in a high-performance game-playing agent.
\newblock In {\em Proceedings of the 16th Annual Meeting of the Special
  Interest Group on Discourse and Dialogue}, pages 77--86, Prague, Czech
  Republic, September. Association for Computational Linguistics.

\bibitem[\protect\citename{Roark}2001]{Roark-2001-td-pars-lm}
Brian Roark.
\newblock 2001.
\newblock Probabilistic top-down parsing and language modeling.
\newblock {\em Computational Linguistics}, 27(2):249--276.

\bibitem[\protect\citename{Sagae \bgroup et al.\egroup
  }2009]{sagae-EtAl-2009-nlu-partial-sr}
Kenji Sagae, Gwen Christian, David DeVault, and David Traum.
\newblock 2009.
\newblock Towards natural language understanding of partial speech recognition
  results in dialogue systems.
\newblock In {\em Proceedings of Human Language Technologies: The 2009 Annual
  Conference of the North American Chapter of the Association for Computational
  Linguistics, Companion Volume: Short Papers}, pages 53--56, Boulder,
  Colorado, June. Association for Computational Linguistics.

\bibitem[\protect\citename{Schlangen and
  Skantze}2009]{schlangen-skantze-2009-incremental-units}
David Schlangen and Gabriel Skantze.
\newblock 2009.
\newblock A general, abstract model of incremental dialogue processing.
\newblock In {\em Proceedings of the 12th Conference of the European Chapter of
  the ACL (EACL 2009)}, pages 710--718, Athens, Greece, March. Association for
  Computational Linguistics.

\bibitem[\protect\citename{Schlangen \bgroup et al.\egroup
  }2009]{schlangenetal-2009-inc-ref-res}
David Schlangen, Timo Baumann, and Michaela Atterer.
\newblock 2009.
\newblock {Incremental Reference Resolution: The Task, Metrics for Evaluation,
  and a Bayesian Filtering Model that is Sensitive to Disfluencies}.
\newblock In {\em Proceedings of SigDial 2009}, London, UK.

\bibitem[\protect\citename{Selfridge \bgroup et al.\egroup
  }2011]{selfridge-EtAl:2011:stab-acc-inc-sr}
Ethan Selfridge, Iker Arizmendi, Peter Heeman, and Jason Williams.
\newblock 2011.
\newblock Stability and accuracy in incremental speech recognition.
\newblock In {\em Proceedings of the SIGDIAL 2011 Conference}, pages 110--119,
  Portland, Oregon, June. Association for Computational Linguistics.

\bibitem[\protect\citename{Skantze and
  Hjalmarsson}2013]{skantze-Hjalmarsson-2013-inc-speech-gen}
Gabriel Skantze and Anna Hjalmarsson.
\newblock 2013.
\newblock Towards incremental speech generation in conversational systems.
\newblock {\em Computer Speech \& Language}, 27(1):243 -- 262.
\newblock Special issue on Paralinguistics in Naturalistic Speech and Language.

\bibitem[\protect\citename{Steedman}2000]{Steedman-2000-syntactic-process}
Mark Steedman.
\newblock 2000.
\newblock {\em The syntactic process}.
\newblock MIT Press, Cambridge, MA, USA.

\bibitem[\protect\citename{Stoness \bgroup et al.\egroup
  }2005]{Stoness-2005-real-world-ref-for-slu}
Scott~C. Stoness, James Allen, Greg Aist, and Mary Swift.
\newblock 2005.
\newblock Using real-world reference to improve spoken language understanding.
\newblock In {\em AAAI Workshop on Spoken Language Understanding}, pages
  38--45.

\bibitem[\protect\citename{Sturt and Lombardo}2005]{sturtLombardo05}
Patrick Sturt and Vincent Lombardo.
\newblock 2005.
\newblock Processing coordinated structures: Incrementality and connectedness.
\newblock {\em Cognitive Science}, 29:291–305.

\bibitem[\protect\citename{Swayamdipta \bgroup et al.\egroup
  }2016]{Swayamdipta2016}
Swabha Swayamdipta, Miguel Ballesteros, Chris Dyer, and Noah~A. Smith.
\newblock 2016.
\newblock Greedy, joint syntactic-semantic parsing with stack lstms.
\newblock In {\em Proceedings of the 20th SIGNLL Conference on Computational
  Natural Language Learning (CoNLL)}, pages 187--197. Association for
  Computational Linguistics, June.

\bibitem[\protect\citename{Tanenhaus \bgroup et al.\egroup
  }1995]{Tanenhaus-1995-int-vis-ling-info}
MK~Tanenhaus, MJ~Spivey-Knowlton, KM~Eberhard, and JC~Sedivy.
\newblock 1995.
\newblock Integration of visual and linguistic information in spoken language
  comprehension.
\newblock {\em Science}, 268(5217):1632--1634.

\bibitem[\protect\citename{Vaswani \bgroup et al.\egroup
  }2017]{vaswani-2017-attention-all-you-need}
Ashish Vaswani, Noam Shazeer, Niki Parmar, Jakob Uszkoreit, Llion Jones,
  Aidan~N Gomez, \L~ukasz Kaiser, and Illia Polosukhin.
\newblock 2017.
\newblock Attention is all you need.
\newblock In I.~Guyon, U.~V. Luxburg, S.~Bengio, H.~Wallach, R.~Fergus,
  S.~Vishwanathan, and R.~Garnett, editors, {\em Advances in Neural Information
  Processing Systems 30}, pages 6000--6010. Curran Associates, Inc.

\bibitem[\protect\citename{von~der Malsburg and
  Vasishth}2011]{malsburg-2011-scanpath-syntactic-reanalysis}
Titus von~der Malsburg and Shravan Vasishth.
\newblock 2011.
\newblock What is the scanpath signature of syntactic reanalysis?
\newblock {\em Journal of Memory and Language}, 65(2):109 -- 127.

\bibitem[\protect\citename{Wahlster}2000]{wahlster-2000-verbmobil}
Wolfgang Wahlster, editor.
\newblock 2000.
\newblock {\em Verbmobil: Foundations of Speech-to-Speech Translation}.
\newblock Springer-Verlag Berlin Heidelberg.

\bibitem[\protect\citename{Wirén}1992]{wiren-1992-thesis}
Mats Wirén.
\newblock 1992.
\newblock {\em Studies in Incremental Natural-Language Analysis}.
\newblock Ph.d. thesis, Linköping University.

\bibitem[\protect\citename{Zhou \bgroup et al.\egroup
  }2016]{zhou-2016-amr-inc-joint-model}
Junsheng Zhou, Feiyu Xu, Hans Uszkoreit, Weiguang QU, Ran Li, and Yanhui Gu.
\newblock 2016.
\newblock Amr parsing with an incremental joint model.
\newblock In {\em Proceedings of the 2016 Conference on Empirical Methods in
  Natural Language Processing}, pages 680--689, Austin, Texas, November.
  Association for Computational Linguistics.

\end{thebibliography}

\end{document}